\documentclass[conference]{IEEEtran}
\usepackage[T1]{fontenc}
\usepackage[english]{babel}
\usepackage{minted}
\usepackage{xcolor}
\definecolor{mygray}{rgb}{0.95, 0.95, 0.95}
\usepackage[table,xcdraw]{xcolor}
\usepackage{makecell} 
\usepackage{float}    
\usepackage{booktabs} 
\usepackage{tabularx} 
\usepackage{float}    
\usepackage{booktabs}

\IEEEoverridecommandlockouts
\usepackage{cite}
\usepackage{amsmath,amssymb,amsfonts}
\usepackage{algorithmic}
\usepackage{graphicx}
\usepackage{subcaption}
\usepackage{textcomp}
\usepackage{xcolor}
\def\BibTeX{{\rm B\kern-.05em{\sc i\kern-.025em b}\kern-.08em
    T\kern-.1667em\lower.7ex\hbox{E}\kern-.125emX}}
\begin{document}

\title{Challenges of Heterogeneity in Big Data: A Comparative Study of Classification in Large-Scale Structured and Unstructured Domains.}
\author{
\IEEEauthorblockN{
González Trigueros Jesús Eduardo, Alonso Sánchez Alejandro,
Muñoz Rivera Emilio, \\
Peñarán Prieto Mariana Jaqueline, 
Mendoza González Camila Natalia
}
\IEEEauthorblockA{
\textit{División de Ingenierías del Campus Irapuato - Salamanca} \\
\textit{Universidad de Guanajuato}\\
Salamanca, Guanajuato, México \\
\{je.gonzaleztrigueros, a.alonsosanchez, e.munozrivera, mj.penaranprieto, cn.mendozagonzalez@ugto.mx\}
}
}

\maketitle

\begin{abstract}
This study analyzes the impact of heterogeneity ("Variety") in Big Data by comparing classification strategies across structured (Epsilon) and unstructured (Rest-Mex, IMDB) domains. A dual methodology was implemented: evolutionary and Bayesian hyperparameter optimization (Genetic Algorithms, Optuna) in Python for numerical data, and distributed processing in Apache Spark for massive textual corpora. The results reveal a "complexity paradox": in high-dimensional spaces, optimized linear models (SVM, Logistic Regression) outperformed deep architectures and Gradient Boosting. Conversely, in text-based domains, the constraints of distributed fine-tuning led to overfitting in complex models, whereas robust feature engineering—specifically Transformer-based embeddings (ROBERTa) and Bayesian Target Encoding—enabled simpler models to generalize effectively. This work provides a unified framework for algorithm selection based on data nature and infrastructure constraints.
\end{abstract} 

\vspace{0.5 cm}

\begin{IEEEkeywords}
Big Data, Heterogeneity, Supervised Classification, Apache Spark, NLP, Hyperparameter Optimization, Feature Engineering.

\end{IEEEkeywords}

\section{Introduction}

The Big Data paradigm has redefined the boundaries of modern computational analysis. Although multiple definitions exist in the literature, academic consensus characterizes this phenomenon primarily through Laney’s original “Three Vs” model: \textit{Volume}, the magnitude of data; \textit{Velocity}, the rate of data generation and transmission; and \textit{Variety}, the diversity of types and sources of information \cite{laney2001}. While Volume and Velocity have been extensively addressed through horizontal scaling and \textit{streaming} processing architectures, Variety (or heterogeneity) remains the most critical methodological challenge for machine learning \cite{gandomi2015}. Heterogeneity implies that there is no “one-size-fits-all” solution; structured data (tabular, numerical) and unstructured data (text, multimedia) require fundamentally different processing pipelines, feature extraction techniques, and classification models. Ignoring these methodological differences may lead to suboptimal models and an inability to capture the underlying semantics in complex data \cite{jagadish2014}. \\

The central objective of this work is to conduct a methodological comparison of the classification strategies required to address massive domains with contrasting natures. We contrast the workflows needed for:

\begin{itemize}
    \item \textbf{Structured Data:} We analyze the Epsilon dataset \cite{sonnenburg2008}, a standard in the literature for binary classification problems with dense, preprocessed numerical features. Here, the methodology focuses on linear separability and the management of mathematically high-dimensional spaces.

    \item \textbf{Unstructured Data:} We examine the IMDB \cite{maas2011} and Rest-Mex \cite{alvarez2021} datasets. These represent the challenges of Natural Language Processing (NLP), where the methodology must span from tokenization and semantic vectorization (\textit{embeddings}) to sentiment classification, addressing issues such as language ambiguity and cultural contextual variability (e.g., Mexican tourism in Rest-Mex).
\end{itemize}

\vspace{0.5 cm}

Through this comparative study, we report the critical divergences in preprocessing and model selection, demonstrating how the nature of the data dictates the architecture of the solution in Big Data environments.

\section{State of the art}

The evolution of the Big Data ecosystem has moved beyond the mere accumulation of massive volumes of information and into a phase of complexity defined by the intrinsic heterogeneity of data \cite{ref7}. Contemporary predictive analytics systems must navigate a fragmented landscape where rigid tabular structures coexist with streams of unstructured text \cite{ref8}. This dichotomy presents fundamental challenges for supervised classification, requiring a comprehensive reassessment of existing methodologies \cite{ref9}.\\

In the domain of tabular data—organized into rows and columns with irregular correlations \cite{ref10} a counterintuitive reality emerges in contrast to the rise of deep learning. Scientific literature consistently confirms that decision tree ensemble algorithms (GBDT), specifically XGBoost, LightGBM, and CatBoost, maintain an undeniable technical dominance in this data modality \cite{ref11}.\\

Unlike neural networks, which are highly sensitive to input scale and require rigorous normalization, GBDT models inherently manage feature heterogeneity without demanding exhaustive preprocessing \cite{ref12}. Across rigorous benchmarks, GBDT not only achieves superior performance metrics compared to tabular deep learning architectures (such as TabNet), but does so with a fraction of the computational and energy cost, and with significantly fewer hyperparameter adjustments \cite{ref8}.\\

A persistent challenge in this setting is the “curse of dimensionality” ($P \gg N$). As dimensionality increases, the volume of the data space grows exponentially, causing Euclidean distance to lose discriminative meaning and allowing unrelated variables to exhibit spurious correlations by chance. In Big Data contexts—where computing covariance matrices for PCA is prohibitive—the current trend favors integrated feature selection techniques (\textit{embedded methods}) within classification algorithms \cite{ref14}.\\

At the opposite extreme of heterogeneity, text processing has undergone a paradigm shift. Historically, classification tasks represented documents as vectors in extremely high-dimensional and highly sparse vocabulary spaces \cite{ref15}. Although linear algorithms such as SVM were effective in these environments, they suffered from substantial loss of semantic and contextual information \cite{ref15}. The modern solution has been the “densification’’ of information through Word Embeddings and pre-trained language models (PLMs) such as BERT. These models transform the sparse space into a denser, lower-dimensional representation where semantic relationships are preserved \cite{ref16}. However, this advancement entails severe scalability costs:

\begin{itemize}
    \item \textbf{Computational Complexity:} The self-attention mechanism in Transformers exhibits a quadratic complexity of $O(L^2)$, making the processing of long documents prohibitively expensive in terms of memory \cite{ref17}.
    \item \textbf{Inference Latency:} Unlike sparse linear models capable of classifying millions of documents on standard CPUs, deep models require costly hardware accelerators, introducing bottlenecks in real-time applications \cite{ref18}.
    \item \textbf{Domain Adaptation:} A classifier trained in one domain (e.g., movie reviews in IMDB) may fail in other contexts due to semantic variability, necessitating expensive adaptation techniques \cite{ref19}.
\end{itemize}

\vspace{0.5 cm}

Despite the significant advances in tabular and textual data classification when treated independently, a critical review of the state of the art reveals a systemic fragmentation in the literature and the absence of unified frameworks to address heterogeneity in Big Data holistically. We identify three principal deficiencies that this study aims to resolve:

\subsection{Methodological Fragmentation and Research Silos}

Current research operates within isolated silos. There exist extensive surveys on Deep Learning for tabular data \cite{ref11} and on Language Models for NLP, yet there is a scarcity of studies that systematically evaluate how to integrate these two worlds into a single, efficient classification pipeline. Most works address heterogeneity by conceptualizing it as problems of “heterogeneous graphs’’ or complex networks \cite{ref21}, overlooking the practical reality of enterprise and industrial databases where massive relational tables coexist with columns of unstructured free text.

\subsection{Lack of Representative Hybrid Benchmarks}

Standard tabular benchmarks systematically exclude text columns or treat them as simple categorical variables, removing semantic richness \cite{ref22}. Text benchmarks focus purely on linguistic content, ignoring associated structured metadata that are vital in real-world applications \cite{ref16}. Recent initiatives such as the AutoML Benchmark have begun to include multimodal tables, but the number of datasets is limited and they do not reach the necessary scale \cite{ref23}.

\subsection{Need for Cost-Effectiveness Evaluation (``Green AI'')}

There is an urgent need to investigate the trade-off between accuracy, sparsity, and resource consumption. Recent studies on the environmental impact of AI suggest reevaluating when Deep Learning is necessary and when classical methods are superior \cite{ref24}. Comparative studies simultaneously evaluating AUC, latency, and computational cost between GBDT and multimodal Transformers are lacking. \\

This paper addresses these gaps through a rigorous comparative analysis contrasting solution architectures for dense numerical data (Epsilon) \cite{sonnenburg2008} against highly sparse and semantic datasets such as Rest-Mex \cite{alvarez2021} and IMDB \cite{maas2011}. Empirical guidelines are provided on how data nature dictates algorithm choice, offering a unified perspective for classification in highly heterogeneous contexts.

\section{Methodology}
The methodological strategy of this study is structured to dissect and quantify the operational divergences required in processing heterogeneous Big Data. Beyond the mere evaluation of algorithms, the experimental design aims to contrast the complete processing workflows (pipelines) dictated by structured and unstructured domains. The approach focuses on identifying how the intrinsic nature of the data—whether a dense numerical matrix or a sparse textual corpus—conditions each stage of machine learning, from dimensionality reduction and feature extraction techniques to the selection of the classification model.

\subsection{Characterization of Data Domains}

The selected datasets represent extreme points on the structuring spectrum, allowing for the isolation and evaluation of the specific challenges of each information morphology: \\

\textit{\textbf{A. Structured Domain (Dense Numerical): Epsilon}} \\
As a standard for evaluating problems of high mathematical density, the Epsilon dataset, introduced in the PASCAL Large Scale Learning Challenge \cite{sonnenburg2008}, is used. This dataset lacks interpretable semantics, forcing the model to rely exclusively on geometric separability in a high-dimensional vector space. Its inclusion allows for measuring raw computational efficiency and the stability of classical algorithms (such as GBDT and SVM) against preprocessed and normalized data, serving as a control baseline for performance in structured environments. \\

\textit{\textbf{B. Unstructured Domain (Global Text): IMDB Movie Reviews}} \\
To address text mining in a standardized context with high sparse dimensionality, the IMDB dataset is employed \cite{maas2011}. It represents the canonical problem of sentiment analysis with binary polarity. Unlike Epsilon, the dimensionality here is dynamic and depends on the vocabulary size. This domain tests the capability of tokenization and vectorization methodologies (embeddings) to transform free text into numerical representations that capture semantic dependencies without incurring prohibitive computational costs. \\

\textit{\textbf{C. Unstructured Domain (Regional and Cultural Text): Rest-Mex}} \\
Introducing a higher level of complexity associated with ``veracity'' and cultural variability, the Rest-Mex dataset from the IberLEF competition is included \cite{alvarez2021}. Focused on tourism recommendations in Mexico, this dataset challenges standard NLP methodologies trained in English. It evaluates the robustness of models against real linguistic ``noise'': regional idioms, morphosyntactic errors, and the class imbalance typical of real user reviews. Its analysis is crucial to understanding how algorithms degrade when moving out of controlled environments and facing the variability of natural human language.

\subsection{Processing Protocol for Structured Domains (Epsilon)}

The experimental workflow for the dense numerical domain was designed using the Epsilon dataset, employing its pre-normalized version to ensure initial numerical stability ($\mu=0, \sigma=1$). The experimental design aims to isolate classifier behavior in response to variability in instance volume and the efficiency of optimization strategies in high-dimensional spaces. The pipeline is structured into five strict sequential stages: \\

\subsubsection{Incremental Stratified Sampling Strategy}

To analyze model scalability and simulate resource-constrained scenarios, the entire dataset was not used in a single run. Instead, five training subsets ($S_i$) were generated through incremental stratified sampling. Sampling percentages were defined as $\rho \in \{2\%, 8\%, 15\%, 30\%, 45\%\}$. \\

Stratified sampling ensures that the original class proportion $P(y)$ is maintained in each $S_i$, avoiding biases induced by undersampling minority classes. This allows performance metrics to be comparable across different data volumes. \\

\subsubsection{Consistent Dimensionality Reduction (PCA)}

Since Epsilon presents a dense feature space with potential high collinearity, Principal Component Analysis (PCA) was applied as a feature extraction technique. To avoid selection bias and ensure that all experiments operate in the same latent vector space, the following projection strategy was adopted: \\

The PCA object was fitted exclusively on the largest-volume subsample ($S_{45\%}$). This allows capturing the most representative global variance structure of the full dataset. The transformation matrix $W$ obtained from $S_{45\%}$ was frozen and applied to project the remaining subsets ($S_{2\%} \dots S_{30\%}$):

\[
X'_{\text{sub}} = X_{\text{sub}} \cdot W_{45\%}
\]

This methodological decision ensures that dimensionality reduction is consistent across all experiments, allowing variations in performance to be attributed to sample size and algorithm, rather than changes in the feature space. \\

\subsubsection{Selection of Classification Architectures}

After dimensionality reduction, the data were split using an 80/20 Hold-Out validation scheme (80\% training, 20\% testing). Five algorithm families were selected, representing different inductive biases and learning paradigms:

\begin{itemize}
    \item \textbf{Support Vector Machines (SVM):} Selected to evaluate efficacy in margin maximization in high-dimensional spaces.
    \item \textbf{Linear Models (Regression):} Used as a low-complexity, highly interpretable baseline.
    \item \textbf{Multilayer Perceptron (MLP):} Representative of connectionist models (Neural Networks), capable of capturing complex non-linearities.
    \item \textbf{Decision Trees (CART):} Evaluated for their ability to perform non-parametric recursive partitioning.
    \item \textbf{Gradient Boosting (LightGBM):} Included as the current state-of-the-art for tabular data, assessing efficiency in memory management and speed using histograms.
\end{itemize} 

\vspace{0.5 cm}

\subsubsection{Hyperparameter Optimization Framework (HPO)}

Recognizing that algorithm performance is highly sensitive to its configuration, a competitive hyperparameter optimization phase was implemented. Four main approaches for exploring the search space $\Theta$ were compared: \\

\textbf{Exhaustive Search:} To determine the optimal configuration for the SVM classifier, a comprehensive hyperparameter search scheme was implemented using \textbf{Grid Search}. This method systematically evaluates all possible combinations within a predefined search space, thus ensuring a complete and deterministic exploration of model behavior. The employed grid included variations in:
\begin{itemize}
    \item Kernel type: \{\texttt{linear}, \texttt{poly}, \texttt{rbf}\};
    \item Regularization: \(C \in \{0.01,\, 0.1,\, 1,\, 5,\, 10,\, 20\}\) (6 values);
    \item Smoothing coefficient: \(\gamma \in \{\text{scale},\, \text{auto},\, 10^{-4},\, 10^{-3},\, 10^{-2},\, 10^{-1}\}\) (6 values);
    \item Polynomial kernel degree: \(degree \in \{2,3,4\}\) (3 values);
    \item Offset parameter for polynomial kernels: \(coef0 \in \{0.0,\, 0.1,\, 0.5\}\) (3 values).
\end{itemize}

\vspace{0.5 cm}

The total number of “valid” combinations depends on the kernel type. Breaking it down:
\[
\begin{aligned}
|\Theta_{\text{rbf}}| &= 6(C) \times 6(\gamma) = 36,\\[3pt]
|\Theta_{\text{linear}}| &= 6(C) \times 1 = 6,\\[3pt]
|\Theta_{\text{poly}}| &= 6(C) \times 6(\gamma) \times 3(degree) \times 3(coef0) = 324.
\end{aligned}
\]

Therefore, the complete grid covered:
\[
|\Theta_{\text{grid}}| = 36 + 6 + 324 = \textbf{366 distinct combinations}.
\]

Each configuration was evaluated using three-fold stratified cross-validation, implying multiple model trainings:
\[
\text{Models trained} = 366 \times 3 = \textbf{1098 trainings}.
\]

Given that the computational cost of SVM with a non-linear kernel grows between \(O(n^2)\) and \(O(n^3)\), and considering the dataset size, the full Grid Search execution required approximately \textbf{12 hours of continuous computation}. This is due to both the sequential nature of the \texttt{libsvm} solver and the lack of GPU acceleration for non-linear kernels in this setup. Finally, after evaluating all configurations, the best combination corresponded to the \textbf{SVM with RBF kernel}, with optimal hyperparameters: \\

\(C = 1\), \(\gamma = 0.0001\), \texttt{class\_weight = balanced}, and \texttt{random\_state = 42}.\\

This configuration achieved the highest average performance across validation folds and was therefore selected as the final model for the evaluation stage. \\


\textbf{Evolutionary Metaheuristics:} As an alternative to overcome the limitations of exhaustive search in high-dimensional and non-differentiable spaces, a global optimization strategy based on \textbf{Genetic Algorithms (GA)} was implemented. This bio-inspired approach enables parallel exploration of multiple regions of the hyperparameter space, avoiding local optima through stochastic operators of selection, crossover, and mutation. \\

The implementation was carried out using the \texttt{DEAP} library on the PyTorch framework. The genome of each individual was designed as a hybrid structure encoding both topological variables (network architecture) and training hyperparameters. The configuration of the evolutionary algorithm is detailed as follows:

\begin{itemize} 
\item \textbf{Individual Encoding:} Each chromosome represents a unique configuration of the MLPTrainer network, composed of 5 genes: 
\begin{itemize} 
    \item \textit{Topology (Hidden Layers):} Variable network structure. 
    \item \textit{Dropout Rate:} Continuous variable for regularization. 
    \item \textit{Learning Rate:} Logarithmic continuous variable. 
    \item \textit{Activation Function:} Categorical variable (e.g., ReLU, Tanh, ELU). 
    \item \textit{Batch Size:} Discrete integer variable. 
\end{itemize}

\vspace{0.5 cm}

\item \textbf{Genetic Operators:}
\begin{itemize}
    \item \textit{Selection:} A \textbf{Tournament Selection} mechanism with size $k=3$ was used, favoring selective pressure toward individuals with higher \textit{fitness} (validation accuracy) while maintaining genetic diversity.
    \item \textit{Crossover:} A modified uniform crossover was applied, where each gene (parameter) from the parents has a probability $P_{cx} = 0.6$ of being exchanged, allowing recombination of successful hyperparameters.
    \item \textit{Mutation:} To introduce variability and avoid premature convergence, random mutation with probability $P_{mut} = 0.3$ was used, altering specific genes with new values sampled from the search space.
\end{itemize}

\item \textbf{Fitness Function:} The performance of each individual was evaluated by training the proposed neural network for a reduced number of epochs, using validation accuracy as the target metric to maximize.
\end{itemize}

\vspace{0.3 cm}

\textit{Computational Cost and Hardware:} Unlike grid search, the iterative nature of the GA allows solutions to be progressively refined. Experiments were run on a mobile workstation equipped with a \textbf{12th Generation Intel Core i7}, 32 GB of RAM, and a \textbf{NVIDIA GeForce RTX 3060} GPU. \\

The evolutionary process and search for the optimal architecture required \textbf{21 hours of continuous computation}. This temporal cost is justified by the complexity of the model found and the depth of exploration in a mixed search space. \\

\textit{Optimal GA Configuration:} Upon algorithm convergence, the best individual exhibited a significantly more robust and deeper architectural configuration than the baseline models. The optimal solution is characterized by high neuron capacity (N=1408) balanced by aggressive regularization:

\[
\begin{aligned} 
\text{Architecture (Layers)} &: [1024 \rightarrow 256 \rightarrow 128] \\ 
\text{Activation Function} &: \texttt{ReLU} \\ 
\text{Learning Rate} &: 1 \times 10^{-5} \\ 
\text{Dropout} &: 0.6132 \\ 
\text{Batch Size} &: 407 
\end{aligned}
\]

This configuration suggests that, for the addressed problem, the model benefits from a wide representation capacity (initial dense layers of 1024 neurons) controlled via a high \textit{dropout} rate ($>60\%$) to prevent overfitting, operating with a non-standard batch size (407) that optimizes memory usage of the specific GPU employed. \\

\textit{Adaptive GA:}

To address the limitations of exhaustive search (Grid Search) in high-dimensional hyperparameter spaces, an Adaptive Genetic Algorithm (AGA) was implemented. Unlike canonical genetic algorithms that use fixed probabilities for genetic operators, this implementation dynamically adjusts the selection probability of crossover and mutation operators. \\

Each individual in the population represents a specific hyperparameter configuration vector $\theta$ for the evaluated algorithm family (e.g., for XGBoost:  
$\theta = \{n\_\text{estimators},\, max\_\text{depth},\, learning\_\text{rate},\, \dots\}$). The fitness was defined as the average F1-Score obtained via $k$-fold cross-validation. Given the size of the Epsilon dataset, validation was limited to $k=2$ with controlled parallelization ($n\_\text{jobs}=2$) to avoid memory saturation and ensure feasibility of iterative execution.\\

The core of the proposal lies in the operator selection mechanism. Instead of a single crossover or mutation method, “toolboxes” of operators were defined:

\begin{itemize}
    \item \textbf{Crossover Operators} ($\Omega_c$): Uniform Crossover and One-Point Crossover.
    \item \textbf{Mutation Operators} ($\Omega_m$): Random Reset Mutation (changing a single gene) and Multi-Point Mutation (changing multiple random genes).
\end{itemize}
\vspace{0.5 cm}

In each generation $t$, a probability weight $w_i(t)$ is assigned to each operator. If an operator produces an offspring with fitness above the parent population average, it is recorded as a “success” ($S_i$). At the end of the generation, weights are updated according to the following learning rule:

\[
w_i(t+1) = w_i(t)\,(1 - \alpha) + \frac{S_i}{N}\,\alpha
\]

Where:

\begin{itemize}
    \item $\alpha$: Learning Rate, set to $0.1$.
    \item $N$: Population size.
    \item $S_i$: Number of successful offspring generated by operator $i$.
\end{itemize}

\vspace{0.5 cm}

Subsequently, the weights are normalized so that $\sum_i w_i = 1$. This mechanism favors exploitation of operators that prove effective for Epsilon’s specific search landscape, reducing the use of destructive operators. The evolutionary cycle was configured with the following control parameters, designed to balance exploration and exploitation under a strict computational budget:

\begin{itemize}
    \item \textbf{Population Size:} 20 individuals.
    \item \textbf{Generations:} 15 iterations.
    \item \textbf{Selection Strategy:} Tournament of size $k=3$.
    \item \textbf{Elitism:} The top 2 individuals (\textit{Top-2}) of each generation are preserved unaltered to ensure monotonic improvement of the best global performance.
    \item \textbf{Mutation Probability:} $P_m = 0.2$ (applied after crossover).
\end{itemize}

\vspace{0.5 cm}

This approach enabled optimization of linear models such as Logistic Regression and LDA, finding robust configurations (e.g., $L1/L2$ regularization and tree depth) that outperformed default parameters. \\

\textbf{Probabilistic Methods:} Simulated annealing played a central role as a navigation mechanism within a highly irregular and non-convex hyperparameter space. Unlike deterministic or purely random methods, simulated annealing provided a dynamic balance between exploration and exploitation by controlling a temperature that progressively decreased during execution. In the initial stages, the high temperature allowed the algorithm to behave exploratorily, frequently accepting transitions to configurations that even worsened the immediate model performance. This property was fundamental to escape local minima, especially in contexts where small variations in hyperparameters produced abrupt performance jumps, as occurs in decision trees and ensembles. \\

During each iteration of the process, the algorithm generated a new state from the current solution by applying random but bounded perturbations to the hyperparameters. The role of simulated annealing consisted not only in evaluating the objective function value associated with that perturbation but also in the probability of accepting it according to an exponential function dependent on temperature. This mechanism allowed for a controlled degree of randomness. Initially, this promoted a broader traversal of regions of the space that a greedy algorithm would immediately discard. As the temperature decreased according to a predetermined schedule, the probability of accepting worse solutions was reduced, transforming the algorithm's behavior toward refined local search around the best-performing configurations. \\

In the case of Decision Trees, simulated annealing was crucial to overcome the difficulty that slight modifications, such as changes in depth or minimum split criteria, generate completely different tree structures. These discontinuities often cause simple optimization methods to get trapped in suboptimal configurations. Simulated annealing enabled exploration of regions where models temporarily worsened but offered paths to more stable configurations, later identified during low-temperature phases. For Random Forest, the role of simulated annealing was even more significant due to the large number of hyperparameters and their nonlinear interactions. The high temperature allowed acceptance of simultaneous variations that increased forest size or altered feature selection, enabling escape from narrow valleys where the model seemed to converge prematurely. The gradual temperature decrease consolidated this broad exploration into a fine search focused on combinations that showed sustained performance improvement. \\

In the case of AdaBoost, where the balance between learning rate and number of estimators is extremely sensitive, simulated annealing acted as a regulatory mechanism preventing the optimization from prematurely favoring seemingly promising but unstable configurations. Acceptance of intermediate solutions, even with lower performance, allowed better understanding of the model dynamics across different learning scales and ensemble sizes. This was essential to achieve configurations that not only reached higher final performance but also exhibited a more stable training curve. \\

\textbf{Bayesian Optimization:} Implemented via Optuna with TPE: To identify the most effective configuration for the LightGBM model, a Bayesian optimization framework was employed using Optuna, leveraging the \textit{Tree-structured Parzen Estimator} (TPE) algorithm. Bayesian optimization is distinguished by explicitly modeling the relationship between hyperparameters and model performance. Instead of exploring the search space exhaustively or randomly, it constructs a probabilistic representation that allows it to select, at each step, hyperparameters with the highest likelihood of improving performance. \\

Unlike deterministic methods such as Grid Search, Bayesian optimization progresses sequentially: each evaluation provides information that updates the probabilistic model, progressively refining its ability to identify promising regions within the search space. This significantly reduces the number of configurations required to achieve good results, which is particularly relevant for complex models such as LightGBM. In Optuna, the TPE algorithm explicitly models two distributions:
\[
l(x) = P(\theta \mid y \leq y^{*}), \qquad 
g(x) = P(\theta \mid y > y^{*}),
\]
where \( \theta \) represents the hyperparameters and \( y^{*} \) is a threshold distinguishing high-performing configurations from poor ones. The selection criterion consists of choosing new hyperparameters that maximize the ratio:
\[
\text{acquisition} = \frac{l(\theta)}{g(\theta)},
\]
thus favoring configurations similar to those that have previously demonstrated good results.

\medskip

The optimization process focused on efficiently exploring a search space defined by the following hyperparameters:

\begin{equation}
\Theta_{\text{Optuna}} = 
\left\{
\begin{aligned}
    &n_{\text{estimators}},\; learning\_rate, \\
    &num\_leaves,\; max\_depth, \\
    &reg\_\alpha,\; reg\_\lambda
\end{aligned}
\right\}
\end{equation}

The explored ranges for each hyperparameter were:

\begin{itemize}
    \item \( n\_estimators \in [50, 120] \)
    \item \( learning\_rate \sim \text{LogUniform}(10^{-5}, 5 \times 10^{-1}) \)
    \item \( num\_leaves \in [20, 300] \)
    \item \( max\_depth \in [2, 10] \)
    \item \( reg\_\alpha \in [0, 1] \)
    \item \( reg\_\lambda \in [0, 1] \)
\end{itemize}

In each iteration, Optuna proposed a configuration \(\theta \in \Theta_{\text{Optuna}}\), evaluated by training a LightGBM model with 20-iteration \textit{early stopping} and using the \texttt{binary\_error} metric. The obtained result was fed back into the TPE probabilistic model, which adjusted its internal distributions to favor areas of the space with higher expected performance.

\medskip

A total of:

\[
n_{\text{trials}} = 15
\]

independent hyperparameter proposals were executed. Thanks to the adaptive nature of the TPE method, this number of trials was sufficient to obtain highly competitive configurations, balancing exploration of new regions and exploitation of those that showed good performance in previous iterations.

\medskip

At the end of the process, Optuna identified the set of hyperparameters that maximized the objective metric. This configuration was selected as the recommended one for LightGBM within the experimental framework. \\

The final stage of the pipeline consisted of training the final models using the optimal configurations \(\theta^*\) found by each HPO strategy. Finally, to maximize generalization and robustness, an Ensemble Metamodel was constructed. This ensemble integrates the predictions of the top-performing classifiers (Top-K) using a weighted voting strategy, aiming to reduce error variance and surpass the performance of any single individual model. \\

\subsection{Advanced Processing Protocol for Unstructured Domains (Rest-Mex)}

Processing unstructured data, specifically in the context of user reviews on tourism platforms, presents a multidimensional challenge that goes beyond simple text cleaning. The Rest-Mex corpus introduces complexities inherent to "wild" natural language: spelling noise, intensive use of emojis as sentiment markers, regional dialectal variations (Mexican slang), and a critical class imbalance that threatens the generalization of any predictive model. To transform this heterogeneous raw material into high-fidelity vector representations suitable for classification, a strict sequential pipeline was designed, composed of three macro-phases: Adaptive Linguistic Normalization and Preprocessing, Synthetic Data Augmentation Strategy, and Neural Semantic Encoding. The logical architecture of each phase is detailed below.\\

\subsubsection{Phase 1: Adaptive Linguistic Normalization and Preprocessing}

The first barrier in textual Big Data analysis is format inconsistency. Texts sourced from the web often contain encoding artifacts and non-informative noise. The preprocessing strategy went beyond standard cleaning, integrating heuristic rules specific to the tourism domain and Mexican Spanish. \\

Before any lexical analysis, a byte-level sanitation stage was applied. Text repair algorithms (ftfy) were used to correct \textit{mojibake} errors (corrupted characters due to incorrect UTF-8/Latin-1 decoding). Subsequently, the entire corpus was standardized under Unicode NFKC normalization (Normalization Form KC), ensuring canonical equivalence of visually identical but computationally distinct characters, unifying the representation space. \\

In modern sentiment analysis, digital pictograms —commonly known as \textit{emojis}— should not be considered noise. These symbols act as high-density semantic condensers, encapsulating affective nuances not always present in plain text. Removing an emoji could lead to the loss of a relevant emotional signal. To preserve this information, a \textit{demojization} process was applied, where each symbol is replaced by a standard textual description. For example, a positive reaction is transformed into the token \texttt{:thumbs\_up:}, and an affection symbol into \texttt{:red\_heart:}. This allows language models to process the affective load of these elements as part of the regular vocabulary. Simultaneously, entities with no semantic value for polarity —such as URLs, emails, or HTML tags— were replaced by specialized masking tokens, e.g., \texttt{<url>}, \texttt{<email>}, or \texttt{<html\_tag>}. This preprocessing reduces text dimensionality and stabilizes the model without altering the linguistic structure of sentences. \\

Mexican Spanish presents a lexical richness that pre-trained models often misinterpret. Polysemous words like \texttt{padre} (meaning ``parent'' or ``excellent'') or colloquial expressions such as \texttt{gacho} (``bad'' or ``unpleasant'') introduce semantic ambiguity. To mitigate this issue, a semantic mapping dictionary for idioms was integrated. Using regular expressions, colloquial terms were replaced with their standardized neutral Spanish equivalents; for instance, \texttt{esta\_chido} is normalized to \texttt{es\_bueno}, and \texttt{muy\_canon} becomes \texttt{muy\_dificil}. This normalization reduces vocabulary variance and aligns the input text with the linguistic distributions used in training large language models. \\

Traditional stopword removal is often destructive in sentiment analysis, as it removes negations ("no", "nunca", "jamás") that invert sentence polarity. An intelligent filtering logic was designed that purges irrelevant connectors and prepositions while strictly protecting a ``whitelist'' of degree modifiers and negations. This ensures that a sentence like "no fue bueno" is not erroneously reduced to "bueno" after preprocessing.\\

\subsubsection{Phase 2: Data Augmentation Strategy}

Exploratory analysis of the dataset revealed a severe class imbalance, typical in online reputation systems, where positive ratings (4 and 5 stars) vastly outnumber negative and neutral ones (1, 2, and 3 stars). Training models under these conditions induces prediction bias toward the majority class. To address this without reducing the dataset size (undersampling), a Synthetic Data Augmentation architecture was deployed, targeting exclusively the minority classes. Sequence-to-sequence neural models (Seq2Seq) were used to generate new training instances that preserve the original semantics while varying syntax and lexicon. \\

This technique leverages asymmetry in translation to produce natural paraphrases. The implemented logical flow was: Español $\rightarrow$ English using MarianMT-based models to translate the original reviews into a pivot language, then English $\rightarrow$ Español using an independent model. This "round-trip" cycle introduces subtle variations in word choice and grammatical structure (e.g., passive-to-active voice changes, synonym substitutions) generated by model uncertainty, creating synthetic training examples that are semantically equivalent but vectorially distinct. \\

To introduce further variability, a paraphrasing stage was integrated using T5 models (Text-to-Text Transfer Transformer) specifically fine-tuned for sentence rewriting. The text (already in English after the first translation stage) was input to the T5 model with a paraphrasing prompt. Beam search was used to select the most coherent and diverse generated sequences. Finally, the paraphrased output was translated back into Español. This technique allows for more aggressive syntactic restructurings than simple back-translation, significantly enriching the feature space of the minority classes. \\

As a complementary regularization mechanism, length modification heuristics were applied: for long reviews, a random percentage of the final text was removed, simulating more concise opinions without losing the initial context; for short reviews, neutral or positive sentiment suffixes (e.g., "... muy recomendable") were appended, forcing the model to focus on the original initial tokens. The outcome of this phase was the creation of a "super-set" of data where negative and neutral classes increased their representation, enabling more robust learning of decision boundaries.\\

\subsubsection{Phase 3: Morphological Abstraction and Neural Semantic Encoding (RoBERTa Embeddings)}

Once the corpus was stabilized and augmented, lexical dimensionality reduction was performed via lemmatization. Using advanced linguistic processing models (based on spaCy), each token was morphologically analyzed to identify its root or lemma. The applied logic transformed all verb inflections, plurals, and gender variations into their canonical form (e.g., "comimos", "comiendo", "comeremos" $\rightarrow$ "comer"). This abstraction is crucial in small to medium-sized corpora, as it consolidates the statistical frequency of key concepts, preventing the model from dispersing attention across multiple variants of the same word. This process was applied differentially, again respecting the "white list" of negations to avoid semantic corruption. \\

The final stage of the pipeline consisted of transitioning from the symbolic space (text) to the continuous vector space (numbers). Traditional statistical approaches (e.g., TF-IDF) were discarded in favor of contextual language models based on the Transformer architecture, capable of capturing polysemy and long-range dependencies. The RoBERTa model (Robustly Optimized BERT Approach) was selected, specifically a version pre-trained on a massive Spanish corpus (bertin-roberta-base-spanish). Unlike standard BERT, RoBERTa removes the next sentence prediction task and employs dynamic masking, resulting in denser and more robust semantic representations. \\

A key methodological innovation was the separate treatment of the "Título" and "Reseña" fields. It was hypothesized that the title contains a high-density summary of sentiment, whereas the review provides narrative context. Independent embeddings were generated for the lemmatized titles and for the review body, and then concatenated horizontally ($[E_{titulo}, E_{reseña}]$), creating a unified feature vector for each instance. This strategy preserves the strong signal from the title, preventing it from being diluted across the length of the review. \\

The final outcome of this pipeline is a dense matrix of semantic features, noise-free, linguistically normalized, and statistically balanced, ready to be ingested by classification algorithms. \\

\subsection{Hybrid Processing Pipeline for the IMDB Dataset}

The IMDB dataset represents the most architecturally complex scenario in this comparative study. Unlike Epsilon (purely numerical) or Rest-Mex (purely textual), IMDB constitutes a hybrid domain that combines tabular metadata, high-cardinality categorical variables, and unstructured free-text fields. The objective of the pipeline was to synthesize these disparate information sources into a unified feature space that maximizes the generalization capacity of predictive models over the continuous target variable: the average rating (\texttt{avg\_vote}). The workflow was articulated in five critical phases, designed to mitigate the curse of dimensionality and prevent data leakage, integrating Bayesian statistical techniques with deep language models. \\

\subsubsection{Phase 1: Critical Cleaning and Data Leakage Prevention}

The integrity of the experiment depends on strictly isolating information available a priori (before the movie release) from information a posteriori (consequences of the movie's success). A feature audit was conducted, resulting in the systematic pruning of variables that introduced noise or constituted predictive "traps." The set of features $X_{drop}$ removed from the input space is defined as:

\begin{align*}
X_{\text{drop}} = \{ & 
\text{imdb\_title\_id}, \text{original\_title}, \text{date\_published}, \\
& \text{reviews\_user}, \text{reviews\_critic}, \text{metascore}, \\
& \text{gross\_income}, \text{budget} 
\}
\end{align*}

Unique Identifiers (\texttt{id}, \texttt{original\_title}) have cardinality equal to $N$ (number of samples), which prevents any statistical generalization and encourages pure memorization by tree-based algorithms. Variables such as the number of critic reviews or box office revenue are highly correlated with success (\texttt{avg\_vote}), but they are consequence metrics, not causal indicators. Including them would invalidate the model for early prediction purposes. \\

\subsubsection{Phase 2: Transformation of Numerical and Categorical Distributions}

Once the dataset was cleaned, disparities in the statistical distributions of the remaining variables were addressed. Exploratory analysis of the variable \texttt{votes} ($v$) revealed a long-tail distribution with extreme positive skew (Skewness $> 0$). The mean was approximately 9,800 votes, while the maximum exceeded 2.2 million, indicating the presence of massive outliers ("blockbuster" movies) that can destabilize the convergence of gradient-based models. To compress the dynamic range and approximate a normal distribution, a shifted logarithmic transformation was applied:

$$v' = \ln(1 + v)$$

The $+1$ shift ensures numerical stability for movies with $v=0$, avoiding mathematical undefined values ($\ln(0) \to -\infty$). \\

The variable \texttt{genre} is not mutually exclusive; a movie can simultaneously belong to "Action," "Crime," and "Drama." Traditional One-Hot Encoding would treat the string "Action, Crime" as a single distinct category, fragmenting information. A Multi-Hot encoding was implemented, decomposing the genre space into a base set $\mathcal{G}$ of 25 unique genres. For each movie $m$, a binary vector $\mathbf{g}^{(m)} \in \{0,1\}^{25}$ was generated such that:
$$\mathbf{g}^{(m)}_j = \begin{cases} 1 & \text{if genre } j \text{ is present in } m \\ 0 & \text{otherwise} \end{cases}$$
This allows the model to learn independent weights for each genre and capture non-linear interactions among them (e.g., the negative correlation between "Horror" and "Musical"). \\

\subsubsection{Phase 3: Target Encoding with Bayesian Smoothing for High Cardinality}

One of the most acute challenges of the IMDB dataset is the massive cardinality of its nominal categorical variables: \texttt{country} (4,837 unique categories) and \texttt{production\_company} (31,104 unique categories). Applying One-Hot Encoding would generate a sparse matrix with over 35,000 columns, introducing noise and prohibitive computational costs. The adopted solution was Target Encoding, which replaces each category with the mean of the target variable ($y$) observed in the training set. To regularize the mean for low-frequency categories, Bayesian Smoothing was applied. The encoded value $\mu_{\text{enc}}$ for a category $c$ is calculated as:

$$\mu_{enc}(c) = \lambda(n_c) \cdot \mu_c + (1 - \lambda(n_c)) \cdot \mu_{global}$$

where $n_c$ is the number of occurrences of category $c$. The weighting factor $\lambda(n_c)$ was modeled using a centered sigmoid function:
$$\lambda(n_c) = \frac{1}{1 + \exp\left(-\frac{n_c - k_{min}}{f}\right)}$$

If $n_c \ll k_{min}$ (rare category), $\lambda \to 0$ and $\mu_{enc} \approx \mu_{global}$. If $n_c \gg k_{min}$ (frequent category), $\lambda \to 1$ and $\mu_{enc} \approx \mu_c$. This allows encoding thousands of categories efficiently into a single continuous dimension without losing statistical robustness. \\

\subsubsection{Phase 4: Reputation Engineering for Multiple Entities}

The columns \texttt{director}, \texttt{writer}, and \texttt{actors} present a dual complexity: they contain lists of multiple entities per cell and a cardinality exceeding 400,000 unique values. A "Historical Reputation" metric was developed to quantify the impact of human capital on the perceived quality of a movie. The lists were flattened to identify each unique individual $p$, and the average votes of all previous movies associated with person $p$ were calculated:
$$R_p = \frac{\sum_{i \in \mathcal{M}_p} y_i}{|\mathcal{M}_p|}$$

For a movie $m$ with a cast $A_m = \{a_1, a_2, \dots, a_k\}$, the input feature \texttt{actor\_rating} was defined as the average of the reputations of its members:
\[
\text{Feature}_{actor}(m) = \frac{1}{k} \sum_{j=1}^{k} \hat{R}_{a_j}
\]
where $\hat{R}_{a_j}$ is $R_{a_j}$ if the actor exists in training, or $\mu_{\text{global}}$ if unknown (Out-of-Vocabulary handling). Additionally, Experience features (\texttt{count}) were generated, defined as $|\mathcal{M}_p|$, allowing the model to distinguish between a single success and a consolidated career. \\

\subsubsection{Phase 5: Deep Semantic Vectorization (Transfer Learning)}

For complex textual features (\texttt{title} and \texttt{description}), frequency-based methods (TF-IDF) are insufficient to capture narrative context or thematic similarity. Transfer Learning was employed using the SentenceTransformer architecture, with the pre-trained \texttt{all-mpnet-base-v2} model, optimized on over one billion sentence pairs for semantic similarity tasks. Let $T$ be the descriptive text of a movie. The encoding function $\Phi$ projects $T$ into a dense high-dimensional vector space $\mathbb{R}^{d}$ (with $d=768$):
$$\mathbf{e}_{desc} = \Phi_{\text{MPNet}}(T)$$

The final feature vector for each instance $m$ was constructed by horizontally concatenating all processed components:
$$\mathbf{X}_{final}^{(m)} = \left[ \mathbf{x}_{num}^{(m)} \parallel \mathbf{x}_{cat\_smooth}^{(m)} \parallel \mathbf{x}_{reputation}^{(m)} \parallel \mathbf{e}_{title}^{(m)} \parallel \mathbf{e}_{desc}^{(m)} \right]$$

This dense matrix integrates the precision of structured data with the semantic richness of unstructured text, providing a robust input for subsequent regression and classification algorithms. \\

\subsection{Computational Deployment Strategy and Validation Protocols}

To conclude the experimental design, it is imperative to establish the critical divergences in the processing infrastructure and evaluation protocols applied to unstructured domains in contrast to the numerical domain. These decisions respond to physical memory limitations and the distributed nature required by the textual volume. \\

Unlike the Epsilon dataset, where Python environment memory constraints necessitated an incremental subsampling strategy (2\% - 45\%), the unstructured datasets (Rest-Mex and IMDB) were processed in their entirety. This massive ingestion capability was enabled by migrating the workflow to a distributed computing ecosystem based on Scala and Apache Spark. The use of Resilient Distributed Datasets (RDDs) and optimized Spark DataFrames allowed vectorizing and classifying the full text corpus without incurring memory overflows (OOM) typical of single-node Python implementations. \\

Given the heterogeneity of sources, validation strategies were adapted to the native structure of each benchmark. In Rest-Mex, the canonical IberLEF challenge split was respected, using the official validation set predefined by the organizers, ensuring direct comparability with existing literature and avoiding selection biases in the partition. For IMDB, lacking a strict official partition for cross-validation, a standard Hold-Out protocol was applied, randomly splitting the total corpus into 80\% for training and 20\% for testing, ensuring a stratified distribution of polarity classes. \\

To maintain methodological consistency with the structured domain, the same five algorithmic families were evaluated: SVM, Linear Models, MLP, Decision Trees, and Ensembles. However, there is a fundamental difference in the scope of optimization. While in Epsilon an intensive hyperparameter search was applied (AGA, Optuna, etc.), in the large-scale unstructured domains (Rest-Mex and IMDB) the models were not subjected to exhaustive optimization. This decision is based on the computational complexity $O(n \cdot d)$ inherent to the high dimensionality of text vectors ($d > 768$ dense dimensions or $>20,000$ sparse) multiplied by the total number of instances ($N_{total}$). Running genetic algorithms or simulated annealing in this distributed environment would be computationally prohibitive. Consequently, for the text domains, this study reports the performance of the best classifiers configured with standard robust parameters, prioritizing the evaluation of pipeline scalability and stability over marginal fine-tuning. \\

\section{Results and Discussion}

The experimental evaluation is structured to contrast the performance of models across the three data domains. In this section, the findings for the structured domain (Epsilon) are reported first, followed by the analysis of the unstructured domains (Rest-Mex and IMDB).

\subsection{Performance in Structured Domain: Epsilon Dataset}

For the Epsilon dataset, characterized by high numerical density and normalized preprocessing, five families of algorithms were evaluated. Given the massive data volume and memory constraints, each model was trained on a specific stratified subset and subjected to a differentiated hyperparameter optimization (HPO) strategy. Table \ref{tab:epsilon} summarizes the performance metrics obtained on the test set, detailing the percentage of data used and the search method applied.

\begin{table}[h!]
\centering
\caption{Performance and Configuration on Epsilon}
\resizebox{\columnwidth}{!}{%
\begin{tabular}{lccc}
\hline
\textbf{Model} & \textbf{Optimization} & \textbf{Subsample (\%)} & \textbf{Test / Train Acc.} \\
\hline
SVM & Grid Search & 8 & 0.8809 / 0.8830 \\
Logistic Regression & AGA & 15 & 0.8800 / 0.8902 \\
MLP & GA & 45 & 0.8752 / 0.8777 \\
LightGBM & Optuna & 30 & 0.8570 / 0.8890 \\
AdaBoost & SA & 2 & 0.8006 / 0.8273 \\
\hline
\end{tabular}%
}
\label{tab:epsilon}
\end{table}

The experiment reveals critical insights regarding the relationship between model complexity, data availability, and efficiency in high-dimensional spaces. The SVM, optimized via Grid Search, achieved the highest overall performance (88.09\%) using only 8\% of the data. Combined with the nearly identical performance of Logistic Regression (88.00\%, using 15\%), this suggests that the Epsilon feature space, after dimensionality reduction (PCA), exhibits high linear separability. The additional complexity of non-linear models provided no significant gains. The MLP required the largest data volume (45\%) to reach competitive accuracy (87.52\%). This confirms that deep neural networks need substantially more data to generalize compared to classical classifiers in tabular domains. \\

\textbf{Ensemble Behavior:} \\

\textit{LightGBM:} Despite being optimized with Optuna and using 30\% of the data, it exhibited slight overfitting (larger gap between Train: 88.9\% and Test: 85.7\%), indicating sensitivity to noise in latent features. \\

\textit{AdaBoost:} The low performance (80.06\%) is attributed to the severe subsampling (2\%) necessary to enable the execution of Simulated Annealing. This highlights the difficulty of optimizing complex iterative ensembles under strict computational constraints. \\

\begin{figure}[h!]
\centering
\includegraphics[width=0.3\textwidth]{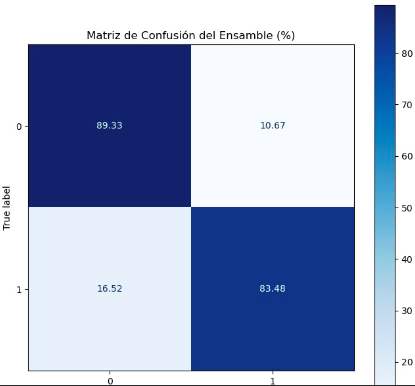}
\caption{Confusion Matrix of the Final Ensemble on Epsilon. The high true positive and true negative rates (main diagonal) demonstrate the generalization capability of the combined system.}
\label{fig:confusion_epsilon}
\end{figure}

To capitalize on individual strengths, a meta-ensemble was constructed using the models reported in the previous table. Figure \ref{fig:confusion_epsilon} shows the normalized confusion matrix, where the system correctly classified 89.33\% of Class 0 and 83.48\% of Class 1. Errors are evenly distributed, validating the robustness of the weighted voting strategy.

\subsection{Performance in Unstructured Domain: Rest-Mex Dataset}

The analysis of the Rest-Mex dataset was conducted under a Multi-task Learning scheme executed on an Apache Spark cluster. Due to the computational infeasibility of performing exhaustive hyperparameter search over the full distributed corpus, the results of the best classifiers with robust standard configurations are reported. Two predictive objectives with divergent semantic complexity levels were evaluated: category prediction (Type), with 3 classes (Restaurant, Hotel, Attraction), and sentiment analysis (Polarity), with 5 classes on a Likert scale (1-5).

\begin{table}[h!]
\centering
\caption{Performance on Rest-Mex}
\begin{tabular}{lcccc}
\hline
\textbf{Task} & \textbf{Model} & \textbf{Acc.} & \textbf{F1} & \textbf{Train / Gap} \\
\hline
Cat (3 Classes) & XGBoost & 0.891 & 0.891 & 0.946 / 5.4\% \\
& RF & 0.738 & 0.733 & 0.742 / <1\% \\
Sent (5 Classes) & RF & 0.663 & 0.542 & 0.510 / N/A \\
& XGBoost & 0.621 & 0.641 & 0.814 / 19.3\% \\
\hline
\end{tabular}
\label{tab:restmex_compacta}
\end{table}

In the classification of establishment type, XGBoost demonstrated clear superiority, achieving 89.12\% accuracy and surpassing Random Forest (73.80\%) by more than 15 percentage points. This task is deterministic: the distinction between a "Hotel" and a "Restaurante" relies on explicit vocabulary (e.g., "habitación", "cama" vs. "mesero", "sabor"). The embeddings effectively captured these keywords, enabling XGBoost to draw precise decision boundaries. The confusion matrix reflects minimal errors across classes. Sentiment analysis posed a greater challenge due to the subjectivity of language and the 5-class granularity. XGBoost showed excellent training performance (81.45\%) but dropped to 62.12\% on the test set, evidencing severe overfitting of nearly 20 points. Without fine-tuning (omitted due to computational limitations), the model memorized noise patterns in the embeddings instead of generalizing the emotional semantics. \\

Random Forest appeared superior in accuracy (66.34\%), but its F1-Score (0.5416) reveals a bias toward the majority class. Inspecting the confusion matrix shows that the model collapsed in predicting 5-star ratings, almost entirely ignoring minority classes. This limits its usefulness for detecting dissatisfied customers, despite the seemingly high Accuracy. The drop in performance from 89\% (Category) to 62–66\% (Sentiment) quantifies the impact of "Variedad" in Big Data. While thematic classification is deterministic, sentiment analysis introduces irony, cultural context, and ambiguity that base models cannot resolve without deeper architectures or hyperparameter tuning, which were infeasible to execute in the distributed infrastructure.

\subsection{Performance in Hybrid Domain: IMDB Dataset}

The IMDB dataset posed the challenge of predicting a continuous variable (average rating from 1 to 10), so regression models were implemented in SparkML. The primary evaluation metric focused on minimizing absolute and squared errors, rather than accuracy. Three architectures were evaluated: Linear Regression (ElasticNet), Random Forest Regressor, and XGBoost Regressor.

\begin{table}[h!]
\centering
\caption{Regression Performance: IMDB}
\begin{tabular}{lccc}
\hline
\textbf{Model} & \textbf{MAE} & \textbf{RMSE} & \textbf{R2} \\
\hline
Linear Regression & 0.915 & 0.698 & 0.446 \\
Random Forest & 0.695 & 0.929 & 0.426 \\
XGBoost & 0.815 & 1.117 & 0.170 \\
\hline
\end{tabular}
\label{tab:imdb_compacta}
\end{table}

The simplest model, Linear Regression ($\alpha=0.5$), achieved the best RMSE (0.698) and highest $R^2$ (0.446), explaining nearly 45\% of the variance in votes. This shows that, after feature engineering (smoothed Target Encoding and embeddings), the relationship between predictors and rating is predominantly linear. Random Forest achieved the best MAE (0.695), indicating that, on average, predictions deviate less than 0.7 points from the actual rating. However, its higher RMSE reflects severe errors for outliers (very high or very low rated movies). Unlike its performance in classification (Rest-Mex), XGBoost Regressor showed a low $R^2$ (0.17) and high RMSE (1.117). The default Spark implementation and the absence of hyperparameter tuning prevented proper handling of the target variable dispersion. In hybrid domains, the complexity of Boosting models can be counterproductive if not finely tuned. Regularized Linear Regression proves to be the most balanced option, demonstrating that good feature engineering can allow simple models to outperform complex "mis-tuned" architectures in Big Data.

\subsection{Global Comparative Analysis}

The experimental evaluation designed to contrast the impact of data heterogeneity on classification and regression algorithms revealed phenomena that challenge the notion that more complex models are always superior in Big Data. When comparing results in the structured domain (Epsilon) versus the unstructured and distributed domains (Rest-Mex and IMDB), a fundamental conclusion emerges: architectural sophistication is irrelevant if it is not aligned with the topology of the feature space and the infrastructure’s capacity to properly optimize the model. \\

In Epsilon, a dense numerical domain, linear and margin-based models dominate over deep networks and tree ensembles. SVM and Logistic Regression (optimized with genetic algorithms) achieved over 88\% accuracy, outperforming MLP and LightGBM. This suggests that the high dimensionality after preprocessing produces a linearly separable space, consistent with Cover’s theorem. The stability of these models, trained on 8\%-15\% subsamples, confirms that a representative and well-optimized sample can outperform a complex model trained on more data but suboptimally configured. \\

In the unstructured domains under Spark, the semantic gap becomes evident. In Rest-Mex, classification of establishment type reached 89.12\% accuracy with XGBoost, while polarity analysis dropped to 62\%. This demonstrates that embeddings (RoBERTa) capture explicit lexicon but cannot resolve ambiguity, irony, or subjectivity without fine-tuning. The inability to perform exhaustive hyperparameter optimization (HPO) in Spark led to almost 20\% overfitting during XGBoost training, a problem that could be controlled in Python using genetic algorithms. \\

In IMDB, the regression task highlighted the importance of linear models with strong feature engineering. Regularized Linear Regression outperformed Random Forest and XGBoost, achieving $R^2 = 0.45$ versus 0.17 for boosting. Smoothed Target Encoding and semantic embeddings effectively "linearized" complex interactions between high-cardinality categorical variables and the continuous target. The collapse of XGBoost underscores the extreme sensitivity of Gradient Boosting algorithms to hyperparameters and the inability to generalize without proper tuning. \\

Finally, the evaluation highlights the risks of relying solely on Accuracy in imbalanced datasets. In Rest-Mex, Random Forest reported high Accuracy for Sentiment but a low F1-Score, biased toward the majority class and failing to detect critical minority classes (negative reviews). There is no algorithmic "silver bullet"; success in heterogeneous Big Data depends on balancing data representation (PCA vs. embeddings), optimization strategy (HPO vs. default parameters), and infrastructure capacity. Investing in feature engineering provides higher returns than merely increasing data volume or algorithmic complexity without regularization.

\section{Conclusions}

The present research has dissected the inherent challenges of the ``Variety'' dimension in the Big Data paradigm, contrasting the effectiveness of classification strategies in structured and unstructured domains. Empirical evidence gathered across the \texttt{Epsilon}, \texttt{Rest-Mex}, and \texttt{IMDB} datasets allows us to assert that algorithmic complexity alone does not guarantee predictive performance; rather, the effectiveness of a machine learning system in large-scale environments critically depends on the alignment between data representation (Feature Engineering) and the optimization capacity enabled by the computational infrastructure. In the structured, high-density numerical domain exemplified by the \texttt{Epsilon} benchmark, it is concluded that the dimensionality of the feature space acts as a natural regularizer. Linear and margin-maximizing models, specifically \texttt{SVM} and Logistic Regression, demonstrated technical superiority over complex non-linear architectures such as \texttt{LightGBM} and Neural Networks. This finding validates that, under appropriate preprocessing (e.g., \texttt{PCA}) and rigorous hyperparameter optimization strategies (Genetic Algorithms), ``simple'' classifiers provide superior generalization capability and resource efficiency. Moreover, training on representative stratified subsamples (8--15\%) does not compromise model convergence, questioning the necessity of processing the entirety of raw data when statistical variance is well contained. \\

On the other hand, the exploration of unstructured and hybrid domains via \texttt{Apache Spark} revealed the limitations of computational brute force in the face of semantic subtlety. While the distributed infrastructure enabled the ingestion of the full \texttt{Rest-Mex} and \texttt{IMDB} corpora, the technical impossibility of performing deep hyperparameter optimization cycles exposed the fragility of ensemble models. The severe overfitting observed in \texttt{XGBoost} for sentiment analysis and regression tasks underscores that Gradient Boosting algorithms require fine-tuning, which is often incompatible with the latency constraints of current distributed systems. In contrast, the success of Linear Regression in predicting \texttt{IMDB} ratings confirms that sophisticated feature engineering—integrating Smoothed Target Encoding and neural embeddings—has a far greater impact on final performance than the choice of classifier. Transforming categorical and textual data into dense, smooth numerical representations allowed a linear model to capture complex patterns that eluded more sophisticated but poorly calibrated tree-based models. \\

In conclusion, this work establishes that addressing heterogeneity in Big Data is not a matter of scale, but of method. The observed performance gap between thematic classification (successful) and sentiment analysis (poor) in \texttt{Rest-Mex} indicates that natural language understanding still requires architectures that transcend classical statistical methods, even when enhanced with modern embeddings like \texttt{RoBERTa}. Future research directions include exploring Early Fusion architectures that integrate multimodal neural networks capable of learning joint representations of text and metadata end-to-end, as well as developing more efficient Distributed Hyperparameter Optimization techniques to democratize fine-tuning of complex models in Big Data clusters without incurring prohibitive costs.

\end{document}